\documentclass{article}

\usepackage[preprint,nonatbib]{neurips_2020}

\usepackage[utf8]{inputenc} 
\usepackage[T1]{fontenc}    
\usepackage{hyperref}       
\usepackage{url}            
\usepackage{booktabs}       
\usepackage{amsfonts}       
\usepackage{nicefrac}       
\usepackage{microtype}      

\usepackage{acronym}
\usepackage{amsmath} 
\usepackage{amsfonts} 
\usepackage{url} 
\usepackage{hyperref}
\usepackage{graphicx}
\usepackage{microtype}
\usepackage[sorting=none]{biblatex}
\usepackage[utf8]{inputenc}
\usepackage{slashbox}
\graphicspath{ {./images/} } 
\usepackage{multirow}
\usepackage{multicol}
\usepackage[dvipsnames]{xcolor}
\usepackage{array}
\usepackage[export]{adjustbox}
\usepackage{caption}
\usepackage{subcaption}
\usepackage{booktabs}
\usepackage{longtable,tabu}
\usepackage{listings}
\newacro{nlp}[\textsc{nlp}]{natural language processing}
\newacro{elmo}[\textsc{ELM}\small{o}]{Embeddings from Language Models}
\newacro{bert}[\textsc{bert}]{Bidirectional Encoder Representations from Transformers}
\newacro{mbert}[{\small{m}}\textsc{bert}]{Multilingual \textsc{bert}}
\newacro{mdistilbert}[{\small{m}}\textsc{d}\small{istil}\textsc{bert}]{Multilingual \textsc{d}\small{istil}\textsc{bert}}
\newacro{distilbert}[{\small{Distil}}\textsc{bert}]{Distil\textsc{bert}}
\newacro{camembert}[Camem\textsc{bert}]{Camem\textsc{bert}}
\newacro{xlmroberta}[\textsc{xlm-r}{\small{o}}\textsc{bert}{\small{a}}]{XLM-RoBERTa}
\newacro{roberta}[\textsc{r}{\small{o}}\textsc{bert}{\small{a}}]{RoBERTa}
\newacro{qa}[\textsc{qa}]{Question Answering}
\newacro{pos}[\textsc{pos}]{Parts-of-Speech}
\newacro{lstm}[\textsc{lstm}]{Long Short-Term Memory}
\newacro{bilstm}[\textsc{b}{\small{i}}\textsc{lstm}]{Bi-directional Long Short-Term Memory}
\newacro{squad}[\textsc{sq}{\small{u}}\textsc{ad}]{Stanford Question Answering Dataset}
\newacro{cnn}[\textsc{cnn}]{Convolutional Neural Network}
\newacro{rnn}[\textsc{rnn}]{Recurrent Neural Network}
\newacro{hmm}[\textsc{hmm}]{Hidden Markov Model}
\newacro{gru}[\textsc{gru}]{Gated Recurrent Units}
\newacro{ner}[\textsc{ner}]{Named-Entity Recognition}

\title{Indic-Transformers: An Analysis of Transformer Language Models for Indian Languages}

\author{
	Kushal Jain$^1$\thanks{\ Equal contribution}, \ Adwait Deshpande$^2$\footnote[1] \ , \ Kumar Shridhar$^1$\ , \ Felix Laumann$^1$\ , \ Ayushman Dash$^1$\\
	$^1$NeuralSpace, $^2$Reverie Language Technologies, \\
	\texttt{\{kushal,kumar,felix,ayushman\}@neuralspace.ai},
	\texttt{adwait.deshpande@reverieinc.com}
}
\date{}
\begin{document}
	\maketitle

	\begin{abstract}
	Language models based on the Transformer architecture \cite{vaswani2017Attention} have achieved state-of-the-art performance on a wide range of \ac{nlp} tasks such as text classification, question-answering, and token classification. However, this performance is usually tested and reported on high-resource languages, like English, French, Spanish, and German. Indian languages, on the other hand, are underrepresented in such benchmarks. Despite some Indian languages being included in training multilingual Transformer models, they have not been the primary focus of such work. In order to evaluate the performance on Indian languages specifically, we analyze these language models through extensive experiments on multiple downstream tasks in Hindi, Bengali, and Telugu language. Here, we compare the efficacy of fine-tuning model parameters of pre-trained models against that of training a language model from scratch. Moreover, we empirically argue against the strict dependency between the dataset size and model performance, but rather encourage task-specific model and method selection. We achieve state-of-the-art performance on Hindi and Bengali languages for text classification task. Finally, we present effective strategies for handling the modeling of Indian languages and we release our model checkpoints for the community : \url{https://huggingface.co/neuralspace-reverie}.
		
	\end{abstract}

\section{Introduction}

    Natural Language Processing (\ac{nlp}) has witnessed a paradigm shift from employing task-specific architectures to fine-tuning the same pre-trained language models for various downstream tasks \cite{Howard_Ruder_2018, Peters_Neumann_Iyyer_Gardner_Clark_Lee_Zettlemoyer_2018}. These language models are trained on large corpora of unlabelled text in an unsupervised manner. With the advent of Transformer-based architectures and various pre-training techniques in the last two years  \cite{Devlin_Chang_Lee_Toutanova_2019, clark2020electra, Yang_Dai_Yang_Carbonell_Salakhutdinov_Le_2020, Liu_Ott_Goyal_Du_Joshi_Chen_Levy_Lewis_Zettlemoyer_Stoyanov_2019}, the state-of-the-art results have improved on various downstream tasks. However, much of this research has been limited to high-resource languages such as English, French, Spanish, and German \cite{Wu_Dredze_2020}. There are 7000+ languages spoken around the world and yet most of the \ac{nlp} systems are largely evaluated on a handful of high-resource languages. Hence, there is a dire need to work on \ac{nlp} beyond resource-rich languages \cite{ruder2020}. Our work contributes to filling this gap, as we focus on \ac{nlp} for Indian languages.
    
     India is a multilingual country with only 10 percent of its population speaking English \cite{india}. There have been some concerted and noteworthy efforts to advance research in Indian languages \cite{arora2020inltk}, but very little work has been done with respect to the recently proposed Transformer-based models \cite{Devlin_Chang_Lee_Toutanova_2019, liu2019RoBERTaa}. Most of the recent works that have tried using these models for their applications have preferred using a multilingual version of \ac{bert}, proposed as \acs{mbert} \cite{Devlin_Chang_Lee_Toutanova_2019}, rather than training monolingual language models from scratch, i.e., without transferring any parameters from other pre-trained models. However, \cite{Wu_Dredze_2020}  shows that \acs{mbert} does not have high quality representations for all languages. For example, \ac{mbert} performs worse than non-\ac{bert} models on a number of downstream tasks for the bottom 30 percent of languages in terms of dataset-size upon which \acs{mbert} is trained~\cite{Wu_Dredze_2020}. Therefore, we see it as important to evaluate language models in the monolingual setting for low-resource languages. In this paper, we evaluate Transformer-based architectures for three Indian languages (Hindi, Bengali, and Telugu). To this end, our contributions are as follows:
    \begin{itemize}
        \item We train four variants of contextual monolingual language models, namely \ac{bert}, \acs{distilbert}, \acs{roberta} and \acs{xlmroberta}, for three Indian languages (Hindi, Bengali and Telugu), which are spoken by more than 60 percent of the Indian population \cite{india}. These languages have different scripts based on an alphasyllabary system \cite{bright1999matter}.
        \item We present an exhaustive analysis of these models by evaluating them on multiple \ac{nlp} tasks: \ac{qa}, \ac{pos} Tagging, and Text Classification. We conduct a wide range of experiments with three different setups that enable us to compare our language model variants with their multilingual counterparts and understand different factors that lead to better results on downstream tasks.  
        \item For each setup, we evaluate models under two training settings which include fine-tuning the entire model and using the language model output as contextual embeddings. We use different layers, namely \ac{lstm}~\cite{Hochreiter_Schmidhuber_2006}, \acs{bilstm}, fully connected and Transformer, on top of these embeddings.
        \item We present a correlation between the available dataset size for training vs the results obtained and empirically demonstrate that using Transformers as feature extractors can lead to competitive results in several downstream tasks. 
        \item We plan to release our language model checkpoints\footnote{\url{https://huggingface.co/neuralspace-reverie}} to aid further research in \ac{nlp} for Indian languages. Finally, we also intend to open-source \textit{mergedQuAD}\footnote{\url{https://github.com/Neural-Space/indic-transformers}}, a combination of XQuAD and MLQA datasets for Hindi, that allows training and comparison of \ac{qa} models for Hindi.
    \end{itemize}
    
    Our paper is structured as follows. In the next Section \ref{background}, we place our research in context with related work in the downstream tasks we examine here. In Section \ref{transformer}, we provide the necessary theoretical background to the various Transformer-based models we anticipate to employ, and the subsequent Sections \ref{experiments} and \ref{results} explain our experimental setup and our results, respectively. We conclude our work with a brief discussion in Section \ref{conclusion}.

\section{Background \& Related Work} \label{background}
In this section, we briefly discuss some relevant work in \ac{nlp} for Indian languages, separately for each downstream task. We start with \ac{qa} before we review existing methods for \ac{pos} tagging and text classification for the three Indian languages we investigate.

\subsection{\acl{qa}}
The \ac{qa} task involves extracting the answer for a given question from a corresponding passage.
Traditional \ac{qa} systems for Indian languages were largely based on multiple moving components, each designed for a specific task. For instance, \cite{Sahu_2012} outlined a pipeline for \ac{qa} systems in Hindi with separate components for query generation/processing, document retrieval, and answer extraction. The question types were limited to four and had separate answer selection algorithms for each. A similar component-based method was proposed by \cite{Banerjee_Naskar_Bandyopadhyay_2014} for the Bengali language. While \cite{Sahu_2012} only used static data for evaluation, \cite{Banerjee_Naskar_Bandyopadhyay_2014} collected data from Wikipedia and annotated it for \ac{qa}. As research significantly progressed in recent years, more advanced algorithms for different components were proposed in these systems. \cite{gupta-etal-2018-mmqa} used \acp{cnn} and \acp{rnn} for the classification of a question, the software Lucene\footnote{\url{https://github.com/apache/lucene-solr}} for the retrieval of the document, and similarity features such as term-coverage and proximity-score for extracting the answers. Similarly, \cite{10.1145/3359988} used a language-independent graph-based algorithm for document retrieval and bidirectional \ac{gru} \cite{DBLP:journals/corr/ChoMGBSB14} for answer extraction in a multilingual setting for English and Hindi. More recently, Transformer-based models have been applied to the task of \ac{qa}, but this has been restricted to fine-tuning of \ac{mbert} \cite{Devlin_Chang_Lee_Toutanova_2019} over translated datasets \cite{gupta2020BERT}. A major issue for all the approaches discussed above is that they use different datasets or translate existing datasets, and hence lack a common benchmark. \ac{qa} in English and other high-resource languages has become more end-to-end largely because of the \acs{squad} dataset \cite{rajpurkar2016SQuADb} and a very clear definition of the task. Lately there have been efforts to make \ac{nlp} more multilingual and many \acs{squad}-like datasets have been curated for low-resource languages \cite{Lewis_Oğuz_Rinott_Riedel_Schwenk_2020, clark2020TyDi, Artetxe_Ruder_Yogatama_2020}.

\subsection{\acs{pos} Tagging}
\ac{pos} tagging for Indian languages has been a challenging task not only due to the lack of annotated data but also because of their rich morphology. Most of the earlier attempts \cite{inproceedings} made to tag such languages relied heavily on morphological rules and linguistic features of a language and did not generalize to other languages. \cite{Shrivastava2008HindiPT} tried to counter this by proposing a simple \ac{pos} tagger based on a \ac{hmm} which achieved reasonable results. Amongst Indian languages, a lot of work has been done on tagging Hindi when compared to other low-resource languages. \cite{FeatureProj} proposed a method to transfer features from a comparatively high-resource Indian language like Hindi using parallel translation corpora which were more easily available. \ac{pos} tagging for code-mixed\footnote{Code-mixing is the mixing of two or more languages or language varieties in speech.} \cite{das} Indian languages was another effort in this direction. It involved tagging code-mixed data in multiple language pairs including English-Bengali, English-Telugu, and English-Hindi, collected from various social media platforms \cite{Pimpale_Patel_2016, Ramesh_Kumar_2016}.  More recently, a lot of surveys have focused on evaluating contextualized embeddings for \ac{pos} tagging across many languages using the Universal Dependencies \cite{nivre2020universal} treebanks. For example,  \cite{Straka_Straková_Hajič_2019} present a comparative analysis of 54 languages on tasks like \ac{pos} tagging, dependency parsing and lemmatization using \ac{elmo} \cite{Peters_Neumann_Iyyer_Gardner_Clark_Lee_Zettlemoyer_2018}, flair \cite{akbik-etal-2018-contextual} and \ac{bert} embeddings. Furthermore, \cite{Pires_Schlinger_Garrette_2019} analyse how \ac{mbert} performs by evaluating it on tasks of \ac{pos} tagging and \ac{ner}. 

\subsection{Text Classification}
Advancements in \ac{nlp} for Indian languages mirrors the progress made for English albeit it has been slower due to the lack of annotated datasets, pre-trained word embeddings, and, more recently, pre-trained language models. Nonetheless, inltk \cite{arora2020inltk} and indic-\ac{nlp} \cite{kunchukuttan2020indicnlpcorpus} have curated datasets, trained embeddings and created benchmarks for classification. Indic-\ac{nlp} provides pre-trained FastText \cite{DBLP:journals/corr/BojanowskiGJM16} word embeddings in multiple Indian languages. They have also curated classification datasets for nine languages providing a common benchmark for future research. Apart from these contributions towards the classification task, we briefly discuss some other notable work.
Similar to tagging code-mixed text, sentiment analysis of code-mixed social media text has also drawn attention \cite{Patra_Das_Das_2018}, involving English-Hindi and English-Bengali datasets that saw various machine learning and deep learning approaches being applied to classification. \cite{Joshi_Goel_Joshi_2020} presented a comparative study of various classification techniques for Hindi. They train \ac{cnn} and \ac{lstm} architectures built upon FastText embeddings. They also evaluate \ac{mbert} sentence embeddings on their translated datasets.

\section{Transformer Language Models} \label{transformer}
The Transformer model by \cite{vaswani2017Attention} uses stacked encoder and decoder layers comprised of multi-headed self-attention layers without using any recurrent or convolutional modules. Despite being a departure from the then state-of-the-art \ac{rnn}-based approaches to translation, the Transformer model showed significant performance improvements on the WMT 2014 translation task\footnote{A detailed task description and the dataset are available at \url{(http://www.statmt.org/wmt14/)}}. Inspired by this use of attention mechanism, the language models described below make use of the encoder layers from this model architecture. 

\subsection{\acl{bert}}
\acs{bert} \cite{Devlin_Chang_Lee_Toutanova_2019}, a bidirectional Transformer model, showed significant improvements in natural language understanding tasks. 
A version of the model was trained on English and another on 104 languages using a masked language modeling approach. The latter version, called \acs{mbert}, was pre-trained on Wikipedia articles. The pre-trained version could be used to further fine-tune the performance on downstream tasks such as \acs{qa} and token classification.

\cite{chanbrandendeepset} observed significant improvements in performance over \acs{mbert} by training a language model from scratch on data in German from multiple sources. We expect to see similar performance improvements in low-resource languages by fine-tuning on monolingual data (Setup \ref{setup2}) as well as training from scratch (Setup \ref{setup3}).

\subsection{\acs{distilbert}}
Although the \ac{bert} language model achieves state-of-the-art performance, it comes at the cost of a large model size having more than 340 million parameters \cite{Devlin_Chang_Lee_Toutanova_2019}. This makes it harder to deploy and use it on resource-constrained devices. \cite{sanh2020DistilBERT} proposed a method to reduce \ac{bert}'s model size by 40\% using distillation. The model, called \acs{distilbert} and consisting of only 66 million parameters runs 60\% faster and retains 97\% language understanding capabilities of the original model. Since we aim to show the effectiveness of Transformer-based language models in a production setting as well, we have included \ac{distilbert} in our experiments.

\subsection{\acs{roberta}}
\cite{liu2019RoBERTaa} performed an analysis of the training procedure of \ac{bert} and showed that \acs{bert}'s performance can be improved by training \acs{bert} on a larger dataset for a longer duration. This model, called \acs{roberta}, shows an improvement of 4-5\% over \acs{bert} on natural language inference and \ac{qa} tasks. Another interesting modification that the authors make is the use of a byte-level BPE (Byte Pair Encoding) tokenizer, instead of a character-level BPE tokenizer used in \acs{bert}, to have the advantage of a universal encoding scheme at the cost of a minor degradation in performance. 
\acs{camembert} \cite{martin2020CamemBERT}, a \acs{roberta} language model trained from scratch on French language data, achieves state-of-the-art performance on downstream tasks using a relatively small web crawled dataset. We aim to verify this in our experiments by including a similar monolingual \acs{roberta} model trained on a web crawled Indian language dataset.

\subsection{\acs{xlmroberta}}
Using over 2 terabytes of web-crawled data, the \acs{xlmroberta} model \cite{conneau2020Unsupervised} achieves state-of-the-art performance on natural language inference and several downstream tasks. More interestingly, their work shows that lower resource languages such as Urdu benefit significantly through cross-lingual training at a large scale. Unlike the language-specific tokenization employed by \ac{mbert}, \acs{xlmroberta} uses Sentencepiece \cite{kudo2018sentencepiece} tokenization on raw text without any degradation in performance.

\section{Experimental Setup} \label{experiments}
All our experiments are run on \ac{bert}-based Transformer language models. We use the Huggingface Transformers library \cite{wolf2019huggingface} for setting up our experiments and for downloading pre-trained models. For each language model, we use three setups: 
\begin{enumerate}
    \item \textbf{Setup A}\label{setup1}:  We directly use the pre-trained multilingual version of the models released along with research papers for \acs{bert}, \acs{distilbert} and \acs{xlmroberta}. \acs{roberta} was only trained on English language data and has no multilingual variant. These models form the baseline for all our experiments. The pre-trained models are downloaded from the Huggingface model hub\footnote{\url{https://huggingface.co/models}}. 
    \item \textbf{Setup B}\label{setup2}: We use the pre-trained models from above and train them further, or fine-tune them, keeping the vocabulary unchanged. We fine-tune separately for Hindi, Bengali, and Telugu using monolingual corpora of each language. With this setup, we want to see the effect that increasing the amount of language data has on the original multilingual model. We train these models on the monolingual data for 5 epochs using approximately 300MB of data for each language. 
    \item \textbf{Setup C}\label{setup3}: As observed in \cite{chanbrandendeepset}, training the German language \ac{bert} model from scratch results in a better performance than \ac{mbert}. For each of the architectures, including \acs{roberta}, we train Hindi, Bengali, and Telugu models on monolingual data. We do not transfer any parameters from either the pre-trained (Setup \ref{setup1}) or the fine-tuned models (Setup \ref{setup2}). The models are trained for 5 epochs on monolingual data for each of the languages. We compare this setup with previous setups to see what impact, if any, does cross-lingual training have on performance.
\end{enumerate}

\subsection{Language Selection}
We choose to evaluate Transformer models on languages from three different regions --- North, West, and South --- of India. These languages also show a great variation in their linguistic features as well as their scripts. This makes them well suited for a comprehensive evaluation of language models.
\begin{enumerate}
    \item \textbf{Hindi}: The Hindi language belongs to the Indo-Aryan family \cite{koul2008modern} and is spoken by around 322 million people in India \cite{india} making it the most widely spoken language in the country. Most speakers of Hindi live in the Northern part of India. Hindi is a verb-final language with its verbs and nouns being inflected for gender and number. It is written in the phonetic Devanagari script with spaces being used to separate words.
    \item \textbf{Bengali}: Spoken by more than 96 million people in the Eastern part of India, Bengali is the next most popularly spoken Indian language after Hindi. It also belongs to the Indo-Aryan family, but unlike Hindi it has no grammatical gender \cite{thompson2012bengali}. Bengali is also written in a phonetic script but is distinct from Devanagari.
    \item \textbf{Telugu}: While Bengali and Hindi belong to the same family of languages, Telugu belongs to a separate family of Dravidian languages. It is spoken by 81 million people, largely in the Indian states of Telangana and Andhra Pradesh in the South. Telugu is an agglutinative language \cite{krishnamurti2003dravidian}, meaning morphemes are combined without a change in form to produce complex words. Its nouns and verbs are inflected for both number and gender. Telugu, like Hindi and Bengali, is written using a phonetic script but with a separate orthography.
    
\end{enumerate}
\subsection{Datasets}
We evaluate the performance of different model architectures on \ac{qa}, \ac{pos} tagging, and Text Classification tasks.
While training our language models from scratch (Setup \ref{setup3}), we use monolingual unlabelled datasets. As described in Setup \ref{setup1} and Setup \ref{setup3}, we train all the language models from scratch to compare the effect of pre-training. The Open Super-large Crawled ALMAnaCH coRpus (OSCAR) dataset \cite{ortiz2020monolingual} is a filtered version of the CommonCrawl dataset and has monolingual corpora for 166 languages. Prior to training, we normalize the OSCAR dataset for Hindi, Bengali, and Telugu using the inltk library.

The \ac{squad} \cite{rajpurkar2016SQuADb}, which is commonly used in the evaluation of Transformer language models, has all its examples in English. Inspired by the original \ac{squad} dataset, there are now a few multilingual datasets available which fulfil the same purpose. The TyDi QA dataset by \cite{clark2020TyDi} covers 11 languages with diverse linguistic features in order to evaluate the performance of language models on languages beyond English. The dataset contains 204K question-answer pairs and is divided into two sets of tasks. The primary tasks include selecting the relevant passage and the minimal answer span from the entire context based on the question. However, we use the dataset for the secondary `Gold Passage' task where the relevant passage is directly provided. The secondary task is similar to \ac{squad} and allows us to readily test existing language models.

Hindi is more widely spoken than Bengali and Telugu, yet, interestingly, does not have a comprehensive \ac{qa} dataset available for training. Both the MLQA \cite{lewis2019mlqa} and the XQuAD \cite{Artetxe_Ruder_Yogatama_2020} datasets contain validation and test sets for Hindi but not for training. The XQuAD dataset contains a machine-translated training dataset that has not been human-verified. We instead combine the training and the evaluation sets from XQuAD and MLQA datasets and split the training and test sets in the ratio 90:10. We refer to this dataset in our paper as \textit{mergedQuAD}. We release the training and test split for \textit{mergedQuAD} dataset for future development.

\ac{pos} tagging occurs at the word level and focuses on the relationship between words within a sentence.
We use open-source treebanks annotated by the Universal Dependencies \cite{nivre2020universal} framework for Hindi and Telugu. We use the UPOS\footnote{\url{https://universaldependencies.org/u/pos/}} tags to evaluate our models on \ac{pos} Tagging.  The treebank for Hindi comprises of 16 tags and the Telugu treebank is tagged using 14 such tags. Finally, for Bengali, we use the tagged sentences which are a part of the Indian corpus in nltk. There are 887 examples overall which we split manually into a training and a validation set. 29 unique tags are used in this dataset.

While \ac{qa} and \ac{pos} tagging look at parts of documents and sentences, text classification considers the entire document before placing it in a category. We have, therefore, included this task for completeness in our evaluation of Transformer-based language models. 
For Hindi, we use the BBC Hindi News Articles\footnote{\url{https://github.com/NirantK/hindi2vec/releases/tag/bbc-hindi-v0.1}} which contains annotated news articles classified into 14 different categories. Since the dataset only has train and validation splits, we evaluate on the validation set only.
We report our results on Bengali on a dataset released by Indic-\ac{nlp} \cite{kunchukuttan2020indicnlpcorpus}. The examples are categorized into 2 different classes. For Telugu\footnote{\url{https://github.com/AI4Bharat/indicnlp\_corpus}} too, we use the classification dataset provided by Indic-\ac{nlp}. Each example in the dataset can be classified into any one of 3 different categories.
We summarize the exact dataset splits for all the languages and tasks in Table \ref{datasets}.

\begin{table*}[hbt!]
\small\centering
\resizebox{\linewidth}{!}{
\begin{tabular}
    { l  c  c  c @{\hspace{0.35cm}}  @{\hspace{0.35cm}} c  c  c @{\hspace{0.35cm}}  @{\hspace{0.35cm}} c  c c }
	\toprule
	& \multicolumn{3}{c @{\hspace{0.5cm}}}{\textsc{Classification}} & \multicolumn{3}{c @{\hspace{0.7cm}}}{\textsc{POS Tagging}} & \multicolumn{3}{c @{\hspace{0.7cm}}}{\textsc{Question Answering}} \\
	
	\cmidrule(l{2pt}r{1cm}){2-4}
	\cmidrule(l{0cm}r{1cm}){5-7}
	\cmidrule(l{-0.2cm}r{1cm}){8-10}
	
	\multirow{-2}{*}[1pt]{\textsc{Dataset}} &   \textsc{Hindi}  &   \textsc{Bengali}    &   \textsc{Telugu} 
	                                        &   \textsc{Hindi}  &   \textsc{Bengali}    &   \textsc{Telugu}
	                                        &   \textsc{Hindi}  &   \textsc{Bengali}    &   \textsc{Telugu}   \\
	\midrule
	\textsc{NAME}  &   BBC    &   Indic-\ac{nlp}    &   Indic-\ac{nlp}   &   UD   &   NLTK    &   UD   &   mergedQuAD   & TyDi-QA  & TyDi-QA    \\
	\midrule
	Train          &   3,468  &   11,200   &   19,199   &  13,304  &  665   &   1051   & 2072   & 2390 &  5563  \\
	Validation            &   867    &   1,400    &   2,399    &  1659    &   222   &   131   & 231    & 113  &  669  \\
	Test           &   -      &   1,400    &   2,399    &  1684    &   -     &   146   & -      & -    &  -   \\
	\bottomrule
\end{tabular}
}
\caption{Detailed break-down of all dataset-splits for the various tasks over which we evaluate our models. Each cell represents the number of examples in each split for every language. For cases where test split is missing, we used validation split for testing and validation was performed on a subset of training data.}
\label{datasets}
\end{table*}

\subsection{Evaluation settings}

For each setup explained above, we perform experiments under two settings:
\begin{enumerate}
    \item \textbf{Freezing}. In this setting, we use Transformer models as feature extractors. Specifically, we freeze the weights of the language model and use its output as feature-based embeddings to train different layers on top. For simplicity we refer to the language model as the base model and the trainable layers that we add on top as the head model. We experiment with multiple head types which include:
    \begin{enumerate}
        \item \textbf{Linear}. This simply involves passing the base model outputs to a linear layer. 
        \item \textbf{Multilinear}. Two linear layers are added on top of the base model.
        \item \textbf{\ac{lstm}}. We keep the configuration of \ac{lstm} layers added on top as flexible as possible by varying all its parameters along with other hyperparameters. As such, the head models can have multiple \ac{lstm} layers, can be bidirectional, or both.
        \item \textbf{Transformer}. A Transformer encoder layer was added above the base model.
    \end{enumerate}
    \item \textbf{Fine-tuning}. In this setting, the entire language model is trained along with a single linear layer on top of the model. \\
\end{enumerate}

\begin{itemize}
    \item We report our results for our text-classification experiments in terms of the accuracy obtained on the validation/test set.  
    \item For \ac{pos} tagging, we report F1 scores for all our experiments. This is calculated using the seqeval package\footnote{\url{https://github.com/chakki-works/seqeval}}. Before feeding the labels and our model's prediction to the metric function, we ensure that we only consider the non-padded elements of the inputs. 
    \item For \ac{qa}, we report the F1 score on the validation set. We use the functions from the SQuAD v2.0 Evalutaion Script\footnote{\url{https://worksheets.codalab.org/rest/bundles/0x6b567e1cf2e041ec80d7098f031c5c9e/contents/blob/}} to calculate this score.
\end{itemize}

\section{Results and Analysis} \label{results}

\subsection{Setup A}

\begin{table*}[hbt!]
\small\centering
\resizebox{\linewidth}{!}{
\begin{tabular}
    { l  c  c  c @{\hspace{0.35cm}}  @{\hspace{0.35cm}} c  c  c @{\hspace{0.35cm}}  @{\hspace{0.35cm}} c  c c }
	\toprule
	& \multicolumn{3}{c @{\hspace{0.5cm}}}{\textsc{Classification (Acc)}} & \multicolumn{3}{c @{\hspace{0.7cm}}}{\textsc{POS Tagging (F1)}} & \multicolumn{3}{c @{\hspace{0.7cm}}}{\textsc{Question Answering (F1)}} \\
	
	\cmidrule(l{0.2cm}r{1cm}){2-4}
	\cmidrule(l{0cm}r{1cm}){5-7}
	\cmidrule(l{0cm}r{0.4cm}){8-10}
	
	\multirow{-2}{*}[1pt]{\textsc{Model}}   &   \textsc{Hindi}  &   \textsc{Bengali}    &   \textsc{Telugu} 
	                                        &   \textsc{Hindi}  &   \textsc{Bengali}    &   \textsc{Telugu}
	                                        &   \textsc{Hindi}  &   \textsc{Bengali}    &   \textsc{Telugu}   \\
	\midrule
	\acs{mdistilbert} (frozen)              &   74.86           &   98.28               &   99.01   
	                                        &   95.83           &   78.45               &   89.98
	                                        &   17.5            &   23.76               &   47.40 \\
	\acs{mdistilbert} (fine-tuned)          &   76.89           &   \textbf{98.58}      &   99.13   
	                                        &   97.02           &   85.78               &   95.67   
	                                        &   42.16           &   54.93               &   76.08 \\
	\cmidrule{1-1}
	\acs{mbert} (frozen)                    &   78.60           &   98.29               &   99.30   
	                                        &   96.60           &   83.38               &   95.30   
	                                        &   33.32           &   27.10               &   47.05 \\
	\acs{mbert} (fine-tuned)                &   78.97           &   97.79               &   99.38   
	                                        &   97.51           &   86.45               &   96.20
	                                        &   56.96           &   72.79               &   \textbf{82.90} \\
    \cmidrule{1-1}
    \acs{xlmroberta} (frozen)               &   \textbf{80.63}  &   98.36               &   \textbf{99.59}
                                            &   97.25           &   87.55               &   95.34
                                            &   19.65           &   33.14               &   53.40 \\
    \acs{xlmroberta} (fine-tuned)           &   79.20           &   98.01               &   99.58
                                            &   \textbf{97.98}  &   \textbf{92.60}      &   \textbf{97.12}
                                            &   \textbf{59.70}  &   \textbf{74.31}      &   81.12 \\
	\bottomrule
\end{tabular}
}
\caption{Summary of the results from experiments based on Setup A, where we evaluate multilingual Transformer models on downstream tasks under two training settings, which are specified for each model variant. The training settings are (a) frozen: we use Transformer models as feature extractors and (b) fine-tuned: we fine-tune the entire model. The \textbf{highlighted} metrics show the highest performance across the models that we compare for every language. A similar tabular structure is used for subsequent tables in this section.}
\label{setupAresutls}
\end{table*}

 A general trend (Table \ref{setupAresutls}) across all three languages is that the results improve as we move from \acs{mdistilbert} to \acs{mbert} and then from \acs{mbert} to \acs{xlmroberta}. Although the \ac{mdistilbert} for Bengali performs better than \acs{xlmroberta}, the improvement is marginal. For Hindi, this trend is most evident. \acs{xlmroberta} improves the validation accuracy on \acs{mdistilbert} approximately by 7\%. Such an increase is less evident for Bengali and Telugu where all model variants achieve high metrics rather easily. For each task and each model variant, we further report results in two different training settings: frozen and fine-tuned. In the former setting, we find that the best performing head model for all experiments (across all tasks and languages) is generally a multilayer \ac{lstm} or \ac{bilstm} which is shown specifically for \ac{pos} tagging in \cite{Devlin_Chang_Lee_Toutanova_2019}. While the current norm in \ac{nlp} literature is to fine-tune the entire language model, the classification results show that freezing gives comparable to better results than fine-tuning. 

 The trend of \acs{xlmroberta} outperforming other variants across all languages is apparent when we compare \ac{pos} tagging results. The performance gain with \acs{xlmroberta} against \acs{mdistilbert} is more evident for Bengali (+8\%) and Telugu (+6\%). For \ac{pos} tagging, the best results are achieved consistently with fine-tuning. Moreover, fine-tuned \acs{xlmroberta} performs better by approximately 6\% for Bengali when compared to the frozen setting.

 \acs{xlmroberta} gives the highest accuracy, except for Telugu(-1.78\%), in the \ac{qa} task as well. The pre-trained variant of \acs{xlmroberta} shows a performance gain in Hindi (+2.74\%) and Bengali (+1.52\%) compared to the next best performing model. 

\subsection{Setup B}

\begin{table*}[hbt!]
\small\centering
\resizebox{\linewidth}{!}{
\begin{tabular}
    { l  c  c  c @{\hspace{0.35cm}}  @{\hspace{0.35cm}} c  c  c @{\hspace{0.35cm}}  @{\hspace{0.35cm}} c  c c }
	\toprule
	& \multicolumn{3}{c @{\hspace{0.5cm}}}{\textsc{Classification (Acc)}} & \multicolumn{3}{c @{\hspace{0.7cm}}}{\textsc{POS Tagging (F1)}} & \multicolumn{3}{c @{\hspace{0.7cm}}}{\textsc{Question Answering (F1)}} \\
	
	\cmidrule(l{0.2cm}r{1cm}){2-4}
	\cmidrule(l{0cm}r{1cm}){5-7}
	\cmidrule(l{0cm}r{0.4cm}){8-10}
	
	\multirow{-2}{*}[1pt]{\textsc{Model}}   &   \textsc{Hindi}  &   \textsc{Bengali}    &   \textsc{Telugu} 
	                                        &   \textsc{Hindi}  &   \textsc{Bengali}    &   \textsc{Telugu}
	                                        &   \textsc{Hindi}  &   \textsc{Bengali}    &   \textsc{Telugu}   \\
	\midrule
	\acs{mbert} (frozen)                    & \textbf{78.18}    &   \textbf{98.08}      &   99.30   
	                                        &   97.40           &   87.97               &   \textbf{96.34}   
	                                        &   15.89           &   28.10               &   43.08  \\
    \acs{mbert} (fine-tuned)                &   77.94           &   98.07               &   \textbf{99.63}   
                                            &\textbf{97.68}     &   \textbf{91.36}      &   95.85   
                                            &\textbf{30.86}     &   \textbf{68.42}      &   \textbf{58.36}   \\
	\bottomrule
\end{tabular}
}
\caption{Summary of results from the experiments based on setup B, where we augment multilingual Transformer models with monolingual data for each language. More specifically, we fine-tune \acs{mbert} with 300 MB of monolingual data.}
\label{setupBresults}
\end{table*}
In this setup, we augment the multilingual Transformer models by fine-tuning the language model on 300 MB of monolingual data for each language.
We did not test this hypothesis on all model variants because the initial results (Table \ref{setupBresults}) of doing so with \acs{mbert} were not encouraging.
Augmenting the \acs{mbert} with data from a single language (Hindi, Bengali, and Telugu in our case) does not lead to massive improvements in performance on downstream tasks, and the results are similar or at best comparable to \acs{mbert}. We believe that one of the possible reasons for this is that the dataset size with which we fine-tune the model is small and that the performances might improve with larger dataset size.
However, for \ac{qa}, there is a considerable drop in performance with this setup when compared to \acs{mbert}. While this drop could be task-dependent, there might be other reasons which need to be studied and analyzed in much more detail. We plan to conduct a more comprehensive analysis pertaining to this experimental setup alone in the future, where we would also include other multilingual models and vary the dataset sizes to understand these behaviours more concretely. 

\subsection{Setup C}

\begin{table*}[hbt!]
\small\centering
\resizebox{\linewidth}{!}{
\begin{tabular}
    { l  c  c  c @{\hspace{0.35cm}}  @{\hspace{0.35cm}} c  c  c @{\hspace{0.35cm}}  @{\hspace{0.35cm}} c  c c }
	\toprule
	& \multicolumn{3}{c @{\hspace{0.5cm}}}{\textsc{Classification (Acc)}} & \multicolumn{3}{c @{\hspace{0.7cm}}}{\textsc{POS Tagging (F1)}} & \multicolumn{3}{c @{\hspace{0.7cm}}}{\textsc{Question Answering (F1)}} \\
	
	\cmidrule(l{0.2cm}r{1cm}){2-4}
	\cmidrule(l{0cm}r{1cm}){5-7}
	\cmidrule(l{0cm}r{0.4cm}){8-10}
	
	\multirow{-2}{*}[1pt]{\textsc{Model}}   &   \textsc{Hindi}  &   \textsc{Bengali}    &   \textsc{Telugu} 
	                                        &   \textsc{Hindi}  &   \textsc{Bengali}    &   \textsc{Telugu}
	                                        &   \textsc{Hindi}  &   \textsc{Bengali}    &   \textsc{Telugu}   \\
	\midrule
	\acs{distilbert} (frozen)               &   79.39           &   98.14               &   99.37
	                                        &   97.63           &   \textbf{92.09}      &   95.67   
	                                        &   13.01           &   24.59               &   32.07  \\
	\acs{distilbert} (fine-tuned)           &\textbf{81.83}     &   98.19               &   99.5
	                                        &   97.93           &   91.05               &   96.14
	                                        &   22.03           &   42.73               &   55.74    \\
	\cmidrule{1-1}
	
	\ac{bert} (frozen)                      &   79.77           & \textbf{98.58}        &   99.42
	                                        &   97.72           &   89.96               &   95.43
	                                        &   12.68           &   19.44               &   34.61 \\
	\ac{bert} (fine-tuned)                  &   81.01           &   98.45               &   99.62    
	                                        &\textbf{98.08}     &   91.14               &   96.14
	                                        &   18.61           &   41.12               &   53.86 \\
    \cmidrule{1-1}
    
    \acs{xlmroberta} (frozen)                     &   74.08           &   98.43               &   99.47         
                                            &   96.56           &   88.80               &   \textbf{96.31} 
                                            &   11.85           &   27.48               &   49.16   \\
    \acs{xlmroberta} (fine-tuned)           &   78.75           &   98.36               &   \textbf{99.50}
                                            &   96.73           &   90.61               &   95.60
                                            &   15.93           &   \textbf{53.03}      &   \textbf{70.80} \\
    \cmidrule{1-1}
    \acs{roberta} (frozen)                        &   78.22           &   97.08               &   97.86   
                                            &   96.13           &   84.28               &   87.15
                                            &   8.82            &   20.95               &   34.61 \\
    \acs{roberta} (fine-tuned)              &   76.82           &   97.72               &   98.88   
                                            &   96.74           &   86.85               &   90.92   
                                            &   \textbf{24.74}  &   39.13               &   58.02 \\
	\bottomrule
\end{tabular}
}
\caption{Summary of results from experiments based on setup C, where we train four variants of contextual monolingual models for three languages.}
\label{setupCresults}
\end{table*}

In setup C, we train monolingual language models from scratch and evaluate their performance on downstream tasks. We observe (Table \ref{setupCresults}) that our model variants for \ac{bert} and \acs{distilbert} perform better than their multilingual counterparts (setup A) across all tasks and languages. While our \acs{xlmroberta} variants post competitive results for Bengali and Telugu, they still fall short of the multilingual model by a small margin.  
\acs{xlmroberta}-Hindi performs poorly than expected on classification, \acs{pos} tagging and \acs{qa} under both the training settings. We believe this behaviour can be attributed to the fact that we put some constraints on our training setup. We trained all our model variants uniformly for the same number of epochs.  Essentially, we conjecture that our \acs{xlmroberta} model for Hindi is under-trained and can be trained more to furnish better results on downstream tasks.  

Another observation that warrants our attention is the relatively poor performance of \acs{roberta} models across all tasks and languages. \acs{roberta} uses a Byte-Level BPE tokenizer, which is known to have poorer morphological alignment with the original text compared to a simple unigram tokenizer, and that this results in a higher performance gap on SQuAD and MNLI tasks \cite{bostrom2020byte}. The authors of \acs{roberta} \cite{liu2019RoBERTaa} choose the Byte-Level BPE tokenizer for higher coverage and notice no drop in performance when it comes to English. However, dealing with morphologically rich languages, as in our case, clearly seems to impact the performance.

Fine-tuning the model for the \acs{qa} tasks results in improvements in Setup C as well (more than 11\% improvement for Bengali and Telugu). Across all setups, there is a marked performance gain when a model is fine-tuned against the dataset compared to when only the classifier head is used. This gain is more marked in the case of \ac{qa} task than it is for \acs{pos} tagging or Text classification. Especially apparent from the results for the \acs{qa} task is that the pre-trained models always perform significantly better than the models trained from scratch. 

\begin{table*}[hbt!]
\small\centering
\resizebox{\linewidth}{!}{
\begin{tabular}
    { l  c  c  c @{\hspace{0.35cm}}  @{\hspace{0.35cm}} c  c  c @{\hspace{0.35cm}}  @{\hspace{0.35cm}} c  c c }
	\toprule
	& \multicolumn{3}{c @{\hspace{0.5cm}}}{\textsc{Classification (Acc)}} & \multicolumn{3}{c @{\hspace{0.7cm}}}{\textsc{POS Tagging (F1)}} & \multicolumn{3}{c @{\hspace{0.7cm}}}{\textsc{Question Answering (F1)}} \\
	
	\cmidrule(l{0.2cm}r{1cm}){2-4}
	\cmidrule(l{0cm}r{1cm}){5-7}
	\cmidrule(l{0.cm}r{0.4cm}){8-10}
	
	\multirow{-2}{*}[1pt]{\textsc{Model}}   &   \textsc{Hindi}  &   \textsc{Bengali}    &   \textsc{Telugu} 
	                                        &   \textsc{Hindi}  &   \textsc{Bengali}    &   \textsc{Telugu}
	                                        &   \textsc{Hindi}  &   \textsc{Bengali}    &   \textsc{Telugu}   \\
	\midrule
	Setup A (frozen)                        &   80.63           &   98.36               &   99.59   
	                                        &   97.25           &   87.55               &   95.34   
	                                        &   33.32           &   33.14               &   53.40 \\
	Setup A (fine-tuned)                    &   79.20           &   98.58               &   99.58   
	                                        &   97.98           &   92.60               &   97.12  
	                                        &   59.70           &   74.31               &   82.9 \\
	\cmidrule{1-1}
	Setup C (frozen)                        &   79.39           &\textbf{98.58}         &   99.47  
	                                        &   97.63           &   92.09               &   96.31   
	                                        &   13.01           &   27.48               &   49.16   \\
	Setup C (fine-tuned)                    &\textbf{81.83}     &   98.45               &   99.50   
	                                        &   98.08           &   91.14               &   96.14   
	                                        &   24.74           &   53.03               &   70.80 \\
    \cmidrule{1-1}
    Indic\ac{bert}       &     -     &   97.14   &   \textbf{99.67}   &     -     &     -     &     -   &   -   &  -    & -   \\
    inltk            &   78.75   &     -     &     -     &     -     &     -     &     -   &   -   &   -   &  -  \\
    \cmidrule{1-1}
    TyDiQA                  &         -  &   -        &     -      &       -    &    -       &    -        &   -   &  \textbf{75.4}    & \textbf{83.3}   \\
    
	\bottomrule
\end{tabular}
}
\caption{Direct comparison between the best results from setup A and setup C and related work. The empty cells denote models were either not evaluated on that particular task/language (e.g., Indic\ac{bert} on \ac{pos} Tagging) or that the model was only evaluated on one specific task (e.g., TyDiQA).}
\label{othermodels}
\end{table*}

Finally, we compare our best results from setup A and setup C with other relevant work in the field. The most recent and relevant work related to ours is by \cite{kakwani2020indicnlpsuite}, who released the Indic\ac{bert} model, a multilingual ALBERT model \cite{lan2020albert} pre-trained on 11 Indian languages. They, however, do not report results for \acs{pos} tagging and \ac{squad}-based \ac{qa} tasks. For Hindi classification, we compare our results with inltk \cite{arora2020inltk} that uses a fine-tuned ULMFiT language model. We compare our results with the baseline results for the TyDi QA Gold Passage task for Bengali and Telugu. We do not have a comparable dataset and baseline for Hindi \acs{qa} tasks and we publish our results with our merged dataset.
We report state-of-the-art results for Hindi and Bengali classification, both attained by our language models in setup C. Even for Telugu, the difference between our model and Indic\ac{bert} is marginal. As for the \acs{qa} task, our results do not match up to the baselines of TyDi QA Gold Passage task. The baseline scores for TyDi QA were calculated by fine-tuning over the entire dataset, and not on individual languages. This strongly suggests that the \acs{qa} tasks benefits a lot from cross-lingual transfer learning.

\begin{figure}[ht]
   
   \includegraphics[width=\linewidth]{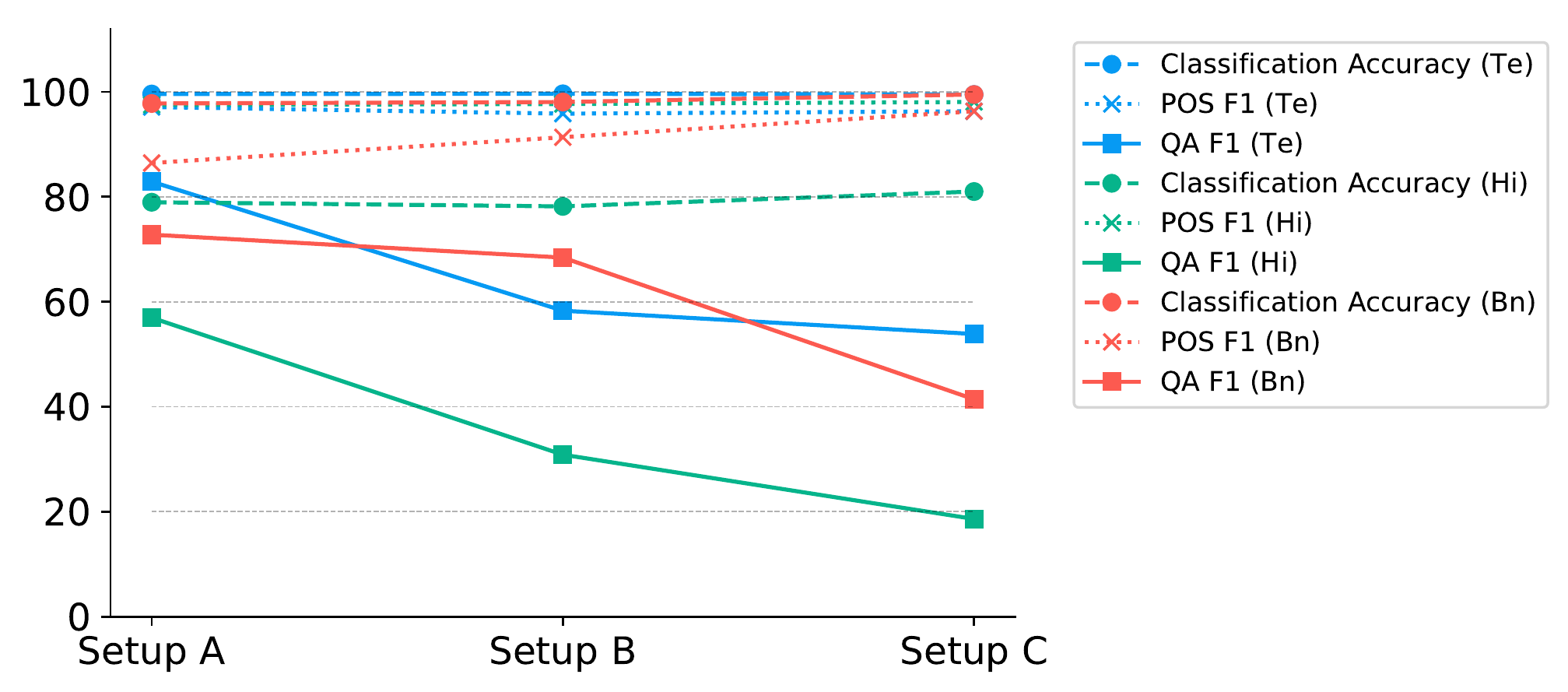} 
    \caption{Relationship between the dataset size that the language model was trained on and its performance on various downstream task. The Y-axis here denotes the generic metric which is F1 score in case of \ac{pos} tagging and \ac{qa} and accuracy in case of text classification.}
\end{figure}

\section{Conclusion and Future Scope} \label{conclusion}
In this work, we have investigated the efficacy of state-of-the-art Transformer language models on languages other than English. We trained 4 variants of monolingual contextual language models with different sizes and capabilities for 3 Indian languages that cover more than 60\% of the country's population. We evaluated our models on three downstream tasks (\acs{pos} Tagging, text classification, and \ac{qa}) under two distinct training settings and across three carefully planned experimental setups. By doing so, we present a highly exhaustive analysis of the performance of Transformer-based models on Indian languages, something which has been lacking in this research space. Our best models reach or improve the state-of-the-art results across multiple tasks and languages. 

One of the most important aspects of our analysis is that we directly compare our trained language models with their existing multilingual counterparts. Upon comparison, our results demonstrate that while monolingual models perform better for some tasks/languages, the improvement attained, if at all, is marginal at best. Our analysis also shows that some variants of Transformer models might be better suited to your needs depending on the available resources, training-dataset size, and downstream task. Furthermore, our experiments also show that competitive results can be achieved by using Transformer models as feature extractors and training different layers on top (\acs{lstm} for best results).  

We observed that a Byte-Level BPE (in \acs{roberta}) affects model performance especially in \acs{qa} tasks. In our follow up work, we aim to explore the impact of tokenizers choice on Indian languages. We notice that the Telugu language model tends to perform well in the \acs{qa} task despite a lower monolingual dataset size (Table \ref{datasize}). This encourages us to explore more on what combination of monolingual dataset size and task level dataset size is sufficient to train models with high accuracy on language inference tasks (an area where there is a lot of room for improvement). 

While there is still a lot of work to do in this area, we outline a few ideas that we hope to pursue in the future. Firstly, we need to create larger corpora for under-resourced languages in order to train more powerful and efficient language models. This work shows that monolingual models trained from the data currently available perform only marginally better than multilingual models. Secondly, we also establish the need of curating annotated datasets for various downstream tasks like \ac{pos} Tagging and \ac{qa} in Indian languages and more importantly, make them easily accessible to the research community. At the same time, we need to create uniformly annotated datasets and benchmarks for Indian languages so that new approaches can be compared easily. \cite{kakwani2020indicnlpsuite} very recently released such a benchmark for Indian languages called the IndicGLUE, which we believe is a step in the right direction.

\printbibliography

\newpage

\section{Appendix}

\appendix
\section{Training details and metrics for Language Modeling} \label{lmdata}
The techniques and setups that we have used for our different language models vary depending on the setup. In our future work we plan to cover in the missing combination of dataset sizes and model architectures for completeness.

\begin{table*}[hbt!]
\small\centering
\resizebox{\linewidth}{!}{
\begin{tabular}
    { l  c  c  c @{\hspace{0.35cm}}  @{\hspace{0.35cm}} c  c  c}
	\toprule
	& \multicolumn{3}{c @{\hspace{0.5cm}}}{\textsc{Setup B}} & \multicolumn{3}{c @{\hspace{0.7cm}}}{\textsc{Setup C}} \\
	
	\cmidrule(l{2pt}r{0.4cm}){2-4}
	\cmidrule(l{0cm}r{0.4cm}){5-7}
	
	\multirow{-2}{*}[1pt]{\textsc{Model}}   &   \textsc{Hindi}  &   \textsc{Bengali}    &   \textsc{Telugu} 
	                                        &   \textsc{Hindi}  &   \textsc{Bengali}    &   \textsc{Telugu}  \\
	\midrule
	\acs{distilbert}                        &  300 MB            &   300 MB             &   300 M
	                                        &   10 GB            &   6 GB              &   2 GB   \\
	\acs{bert}                              &  300 MB            &   300 MB             &   300 M
	                                        &   3 GB            &   3.1 GB              &   1.6 GB    \\
    \acs{roberta}                           &   -               &   -                   &   -
	                                        &   10 GB           &   6 GB               &   2 GB    \\ 
	\acs{xlmroberta}                        &   300 MB          &   300 MB              &   300 MB
	                                        &   3 GB            &   3.1 GB              &   1.6 GB    \\
    
	\bottomrule
\end{tabular}
}
\caption{A comparison of the dataset sizes used for training different language models}
\label{datasize}
\end{table*}

\section{Training metrics for \acs{pos} Tagging}
In this section we show some of the plots for our experiments for each language. These plots clearly show the different head layers that we train as part of our training setting described in the paper. We used Weights and Biases \cite{wandb} to run extensive hyperparameter search for our models. The plots in this section are for Distil\ac{bert} in setup C.
\subsection{Telugu}
\includegraphics[width=0.33\linewidth]{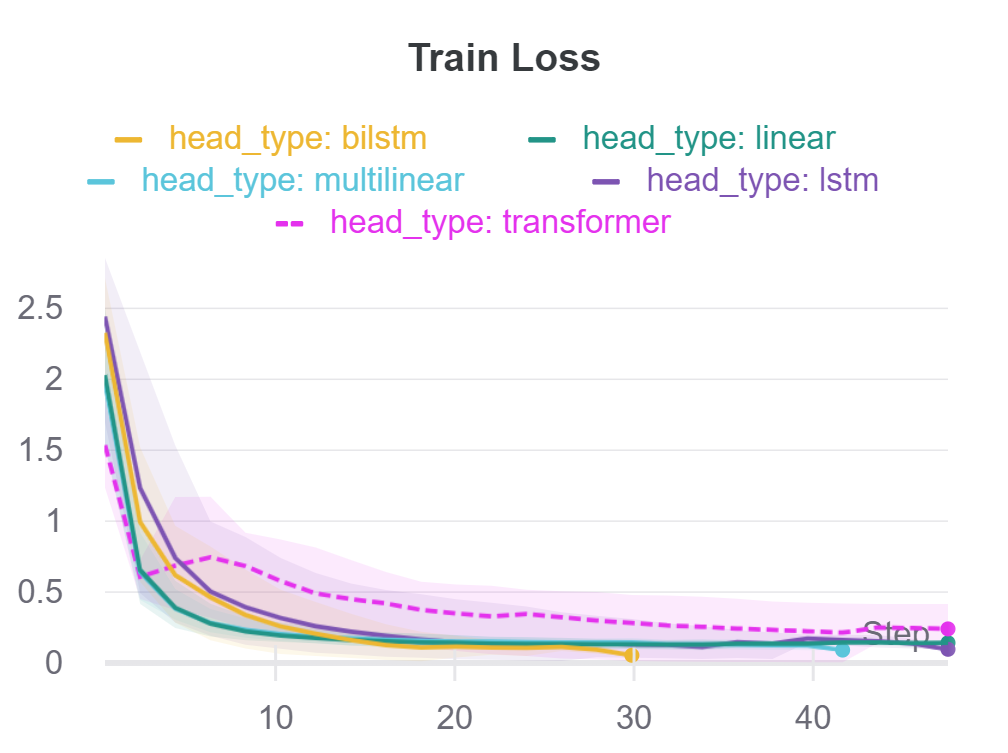}
\includegraphics[width=0.33\linewidth]{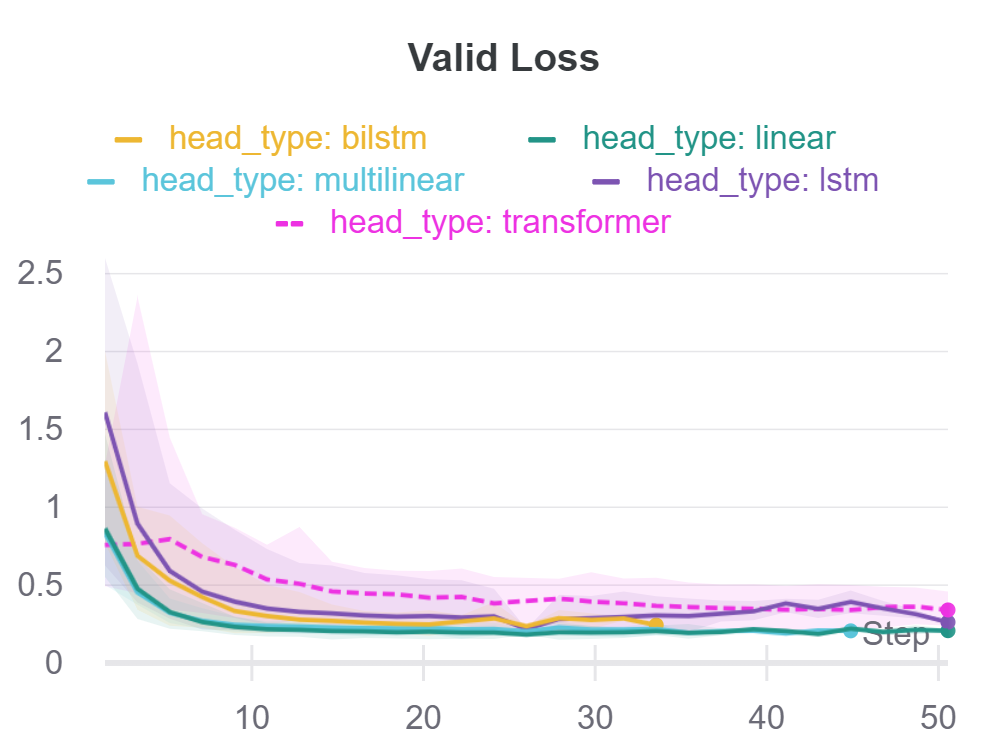}
\includegraphics[width=0.33\linewidth]{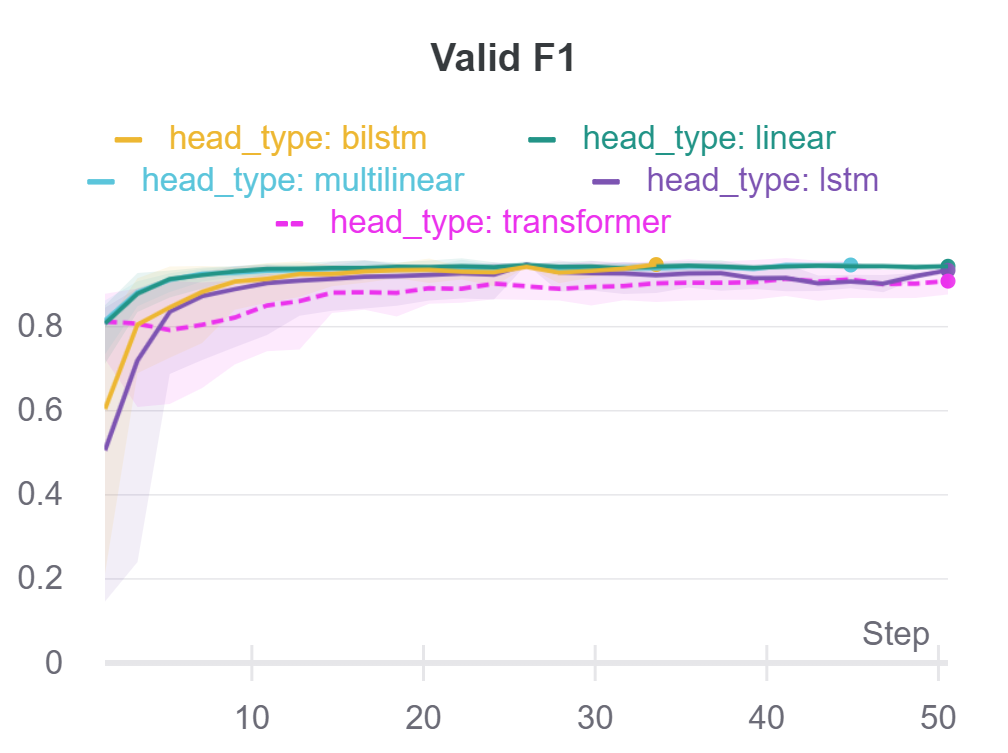}

\subsection{Bengali}
\includegraphics[width=0.33\linewidth]{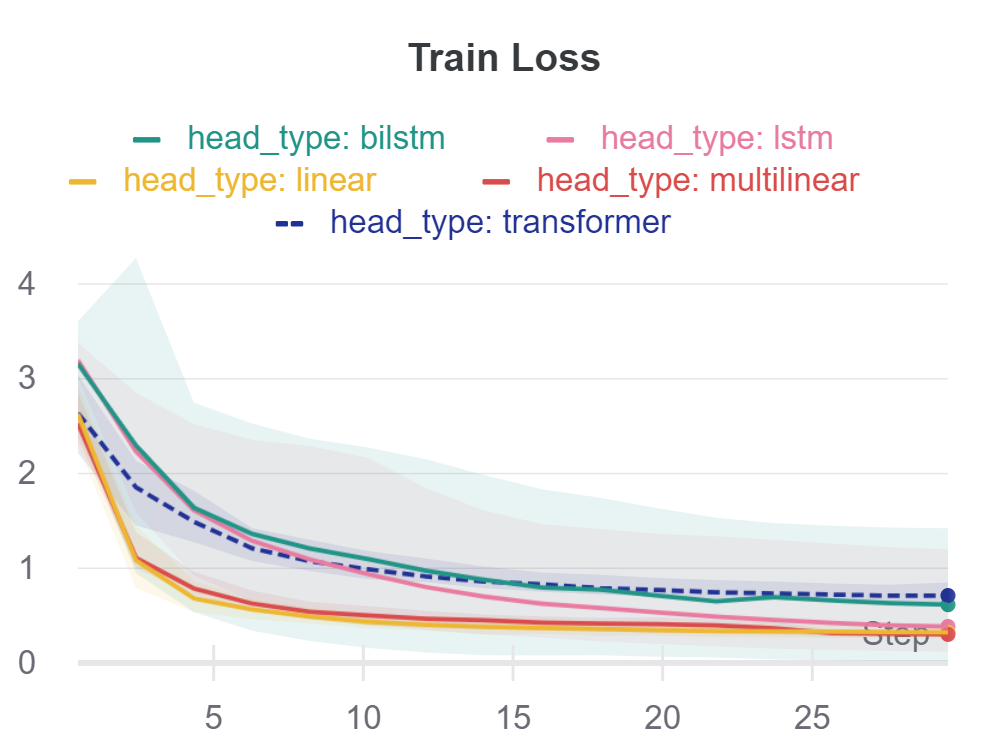}
\includegraphics[width=0.33\linewidth]{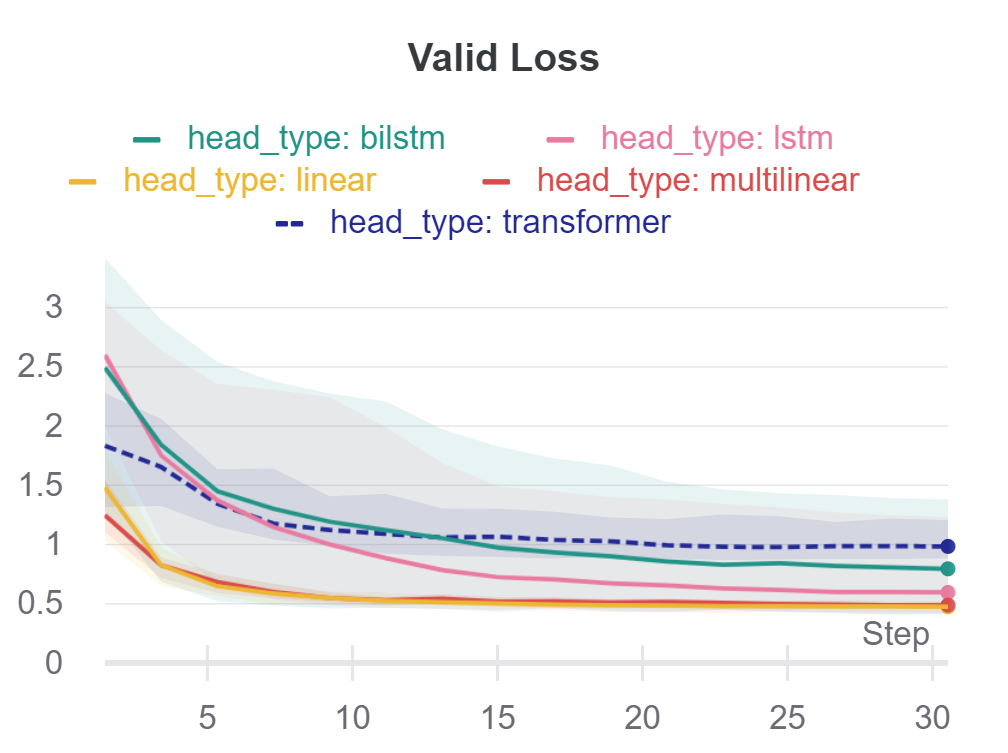}
\includegraphics[width=0.33\linewidth]{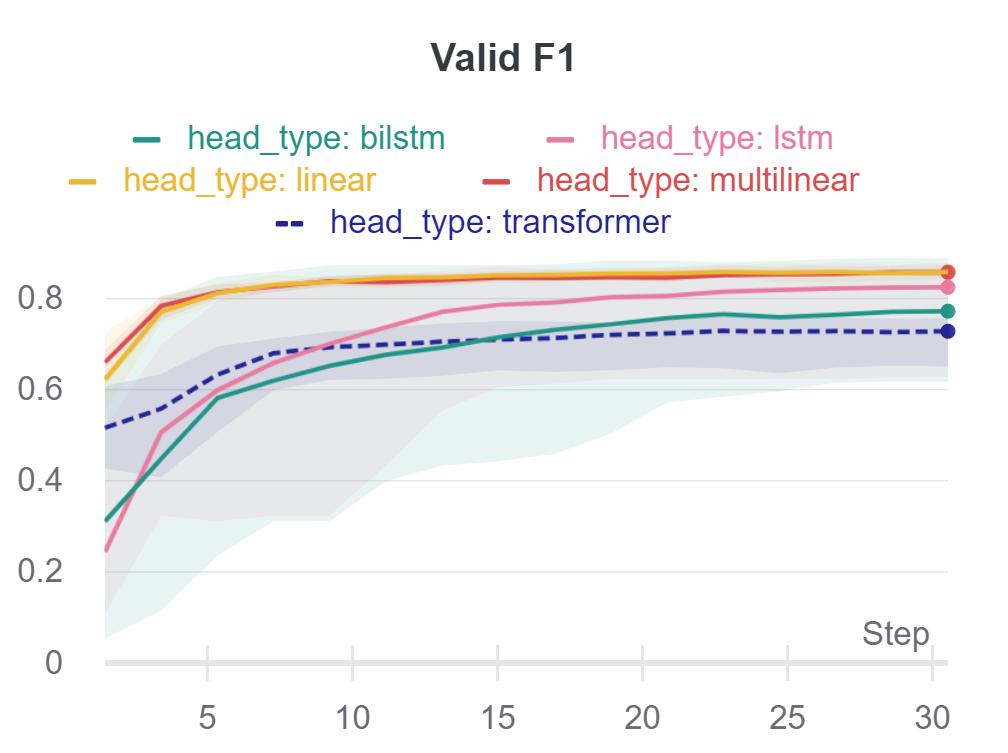}

\subsection{Hindi}
\includegraphics[width=0.33\linewidth]{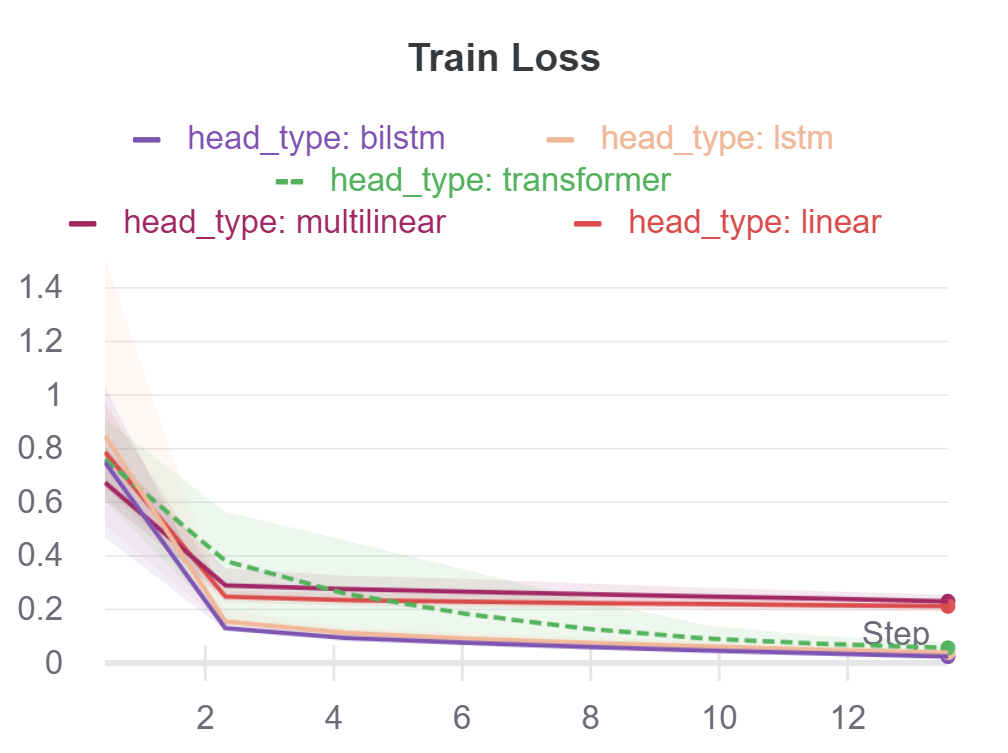}
\includegraphics[width=0.33\linewidth]{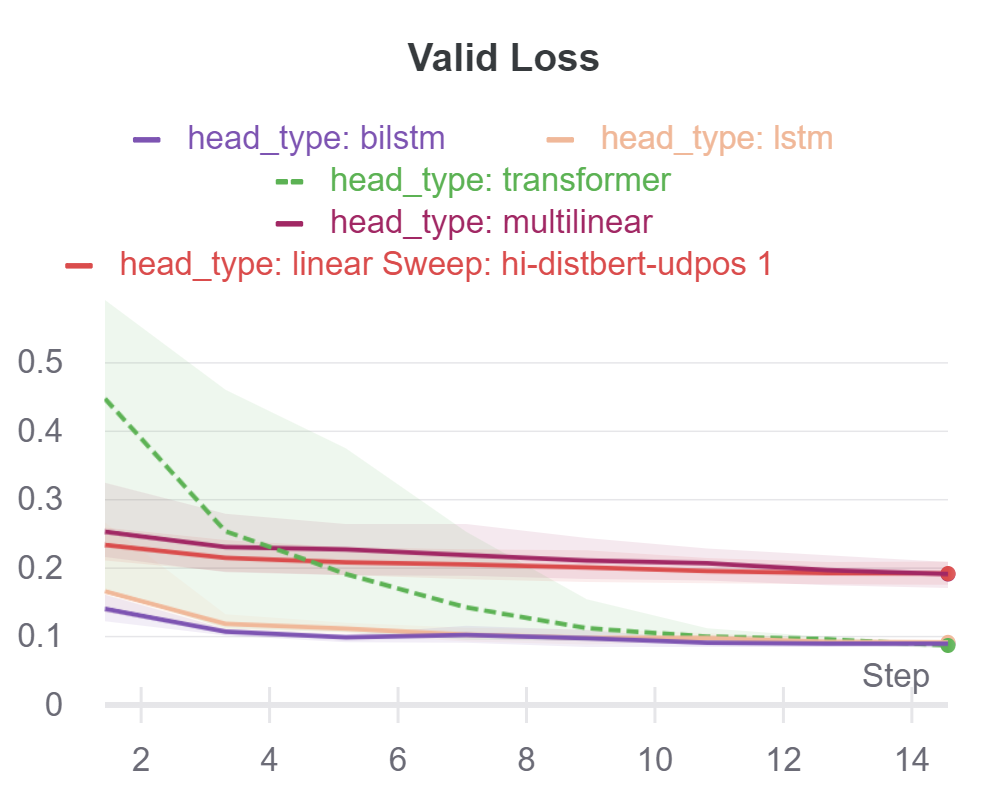}
\includegraphics[width=0.33\linewidth]{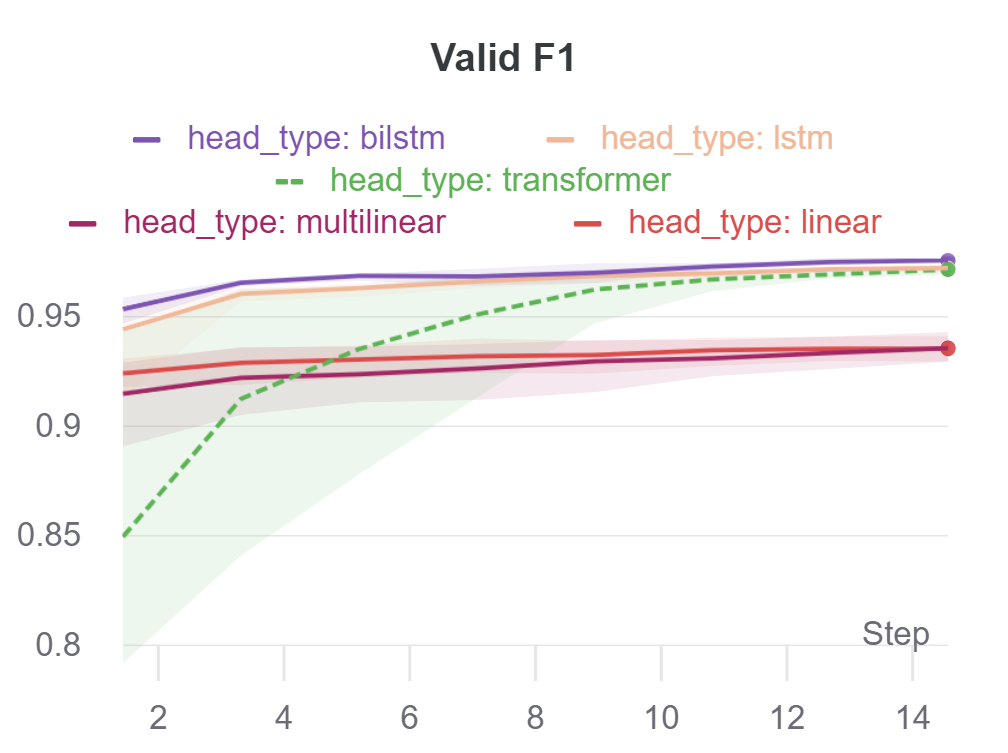}

\section{Training metrics for Text Classification}
This section shows similar graphs for text classification. Here we have taken plots for \ac{mdistilbert} from setup A to demonstrate our experimental setups.
\subsection{Telugu}
\includegraphics[width=0.33\linewidth]{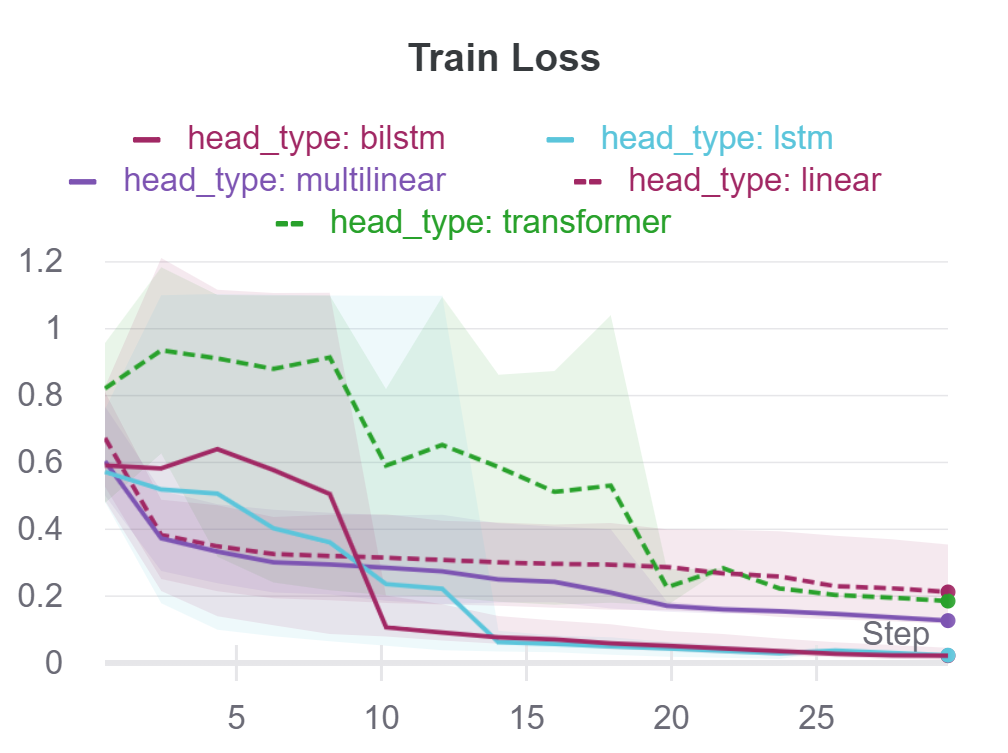}
\includegraphics[width=0.33\linewidth]{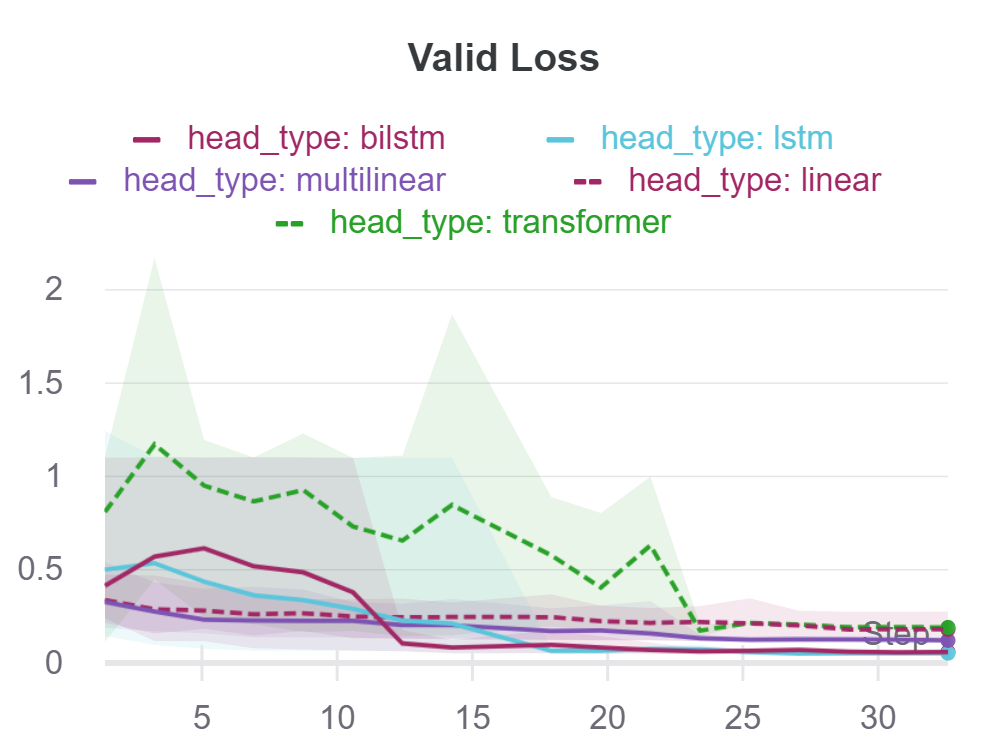}
\includegraphics[width=0.33\linewidth]{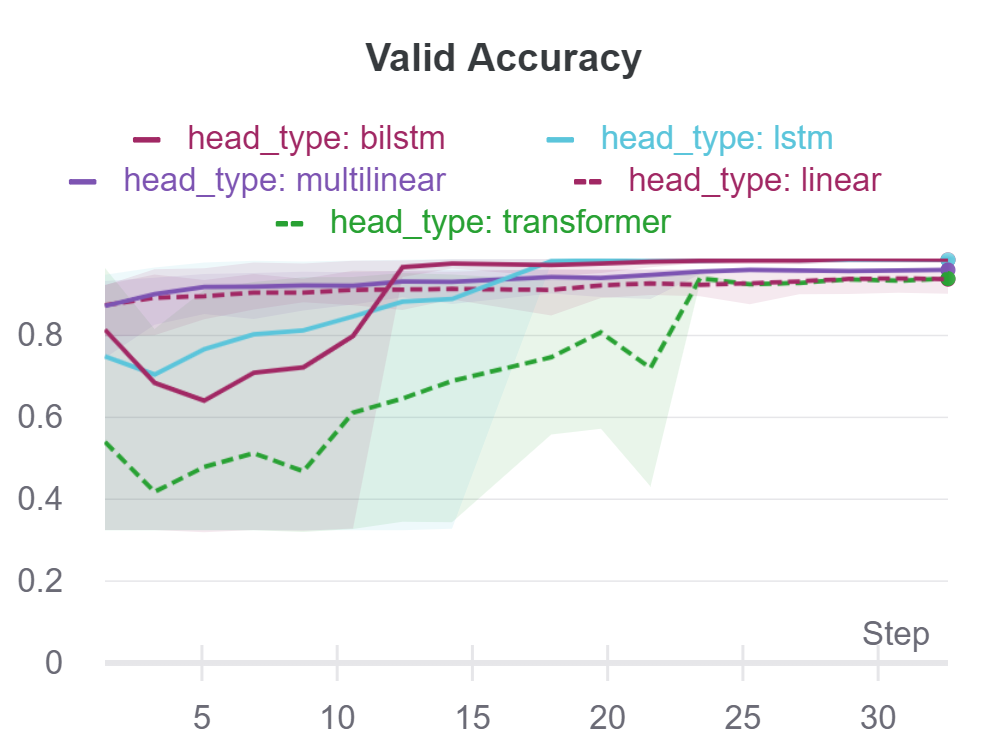}

\subsection{Bengali}
\includegraphics[width=0.33\linewidth]{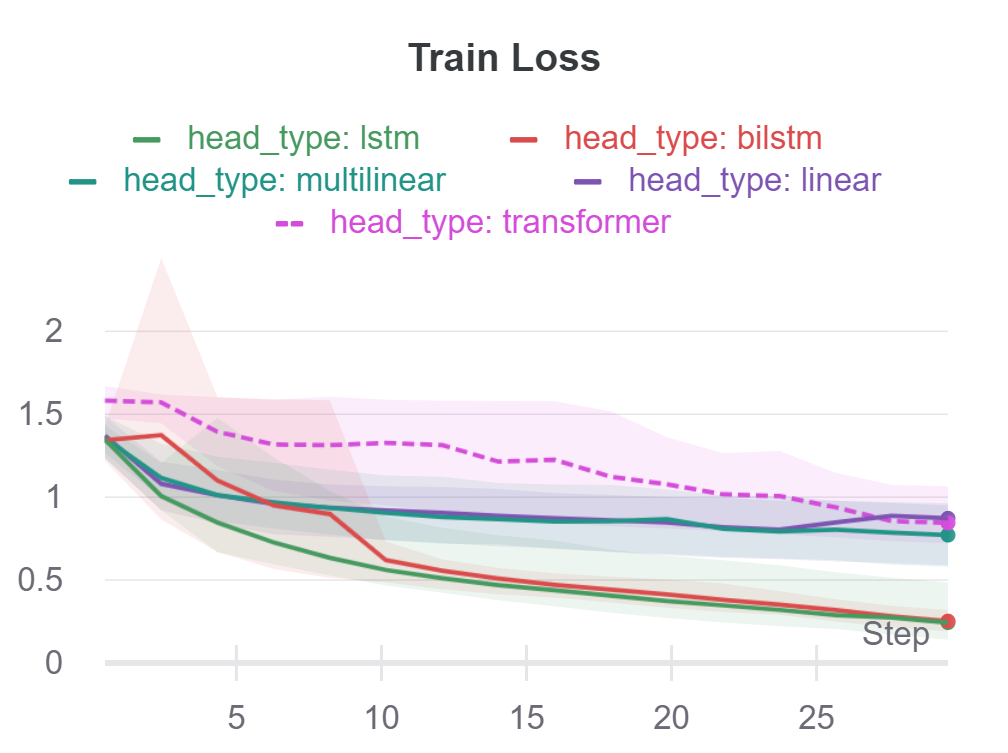}
\includegraphics[width=0.33\linewidth]{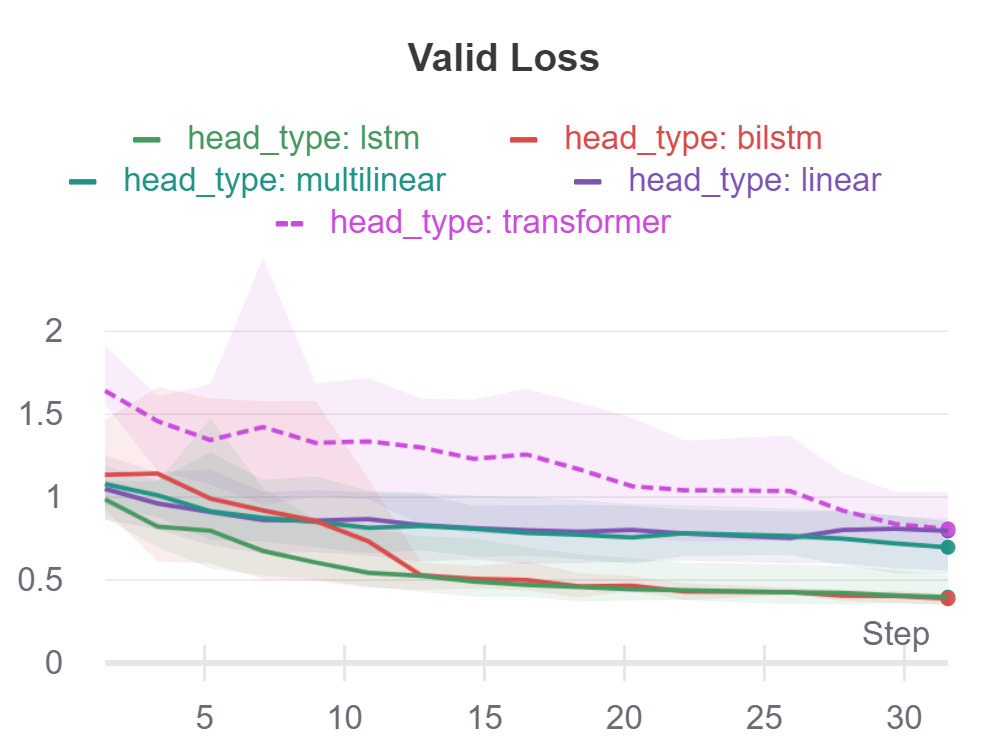}
\includegraphics[width=0.33\linewidth]{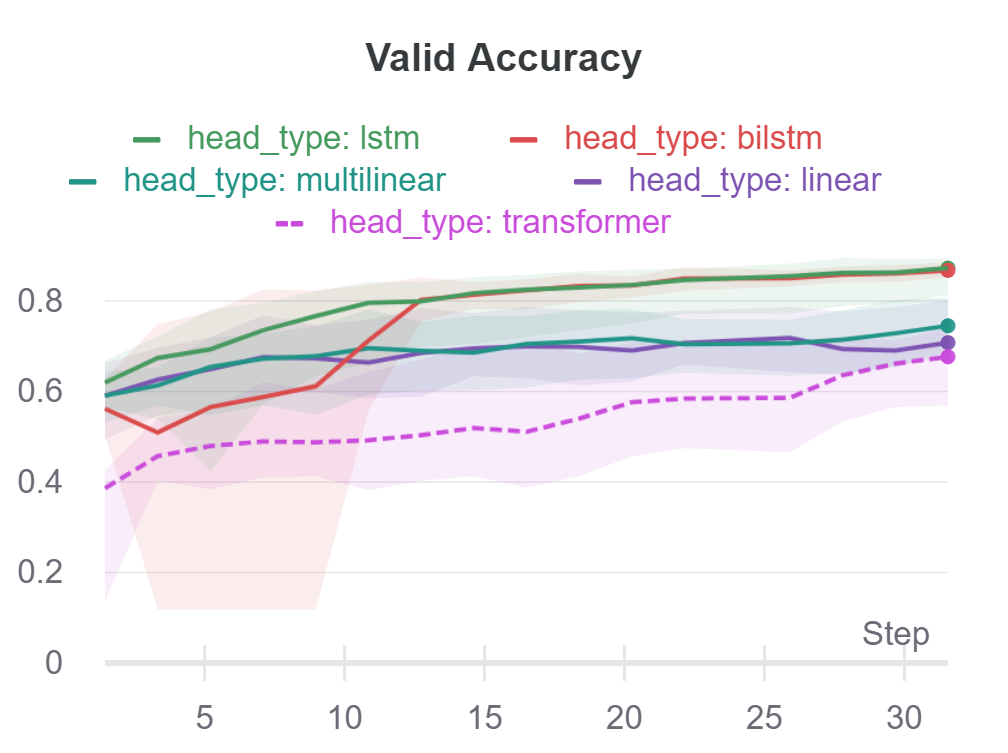}

\subsection{Hindi}
\includegraphics[width=0.33\linewidth]{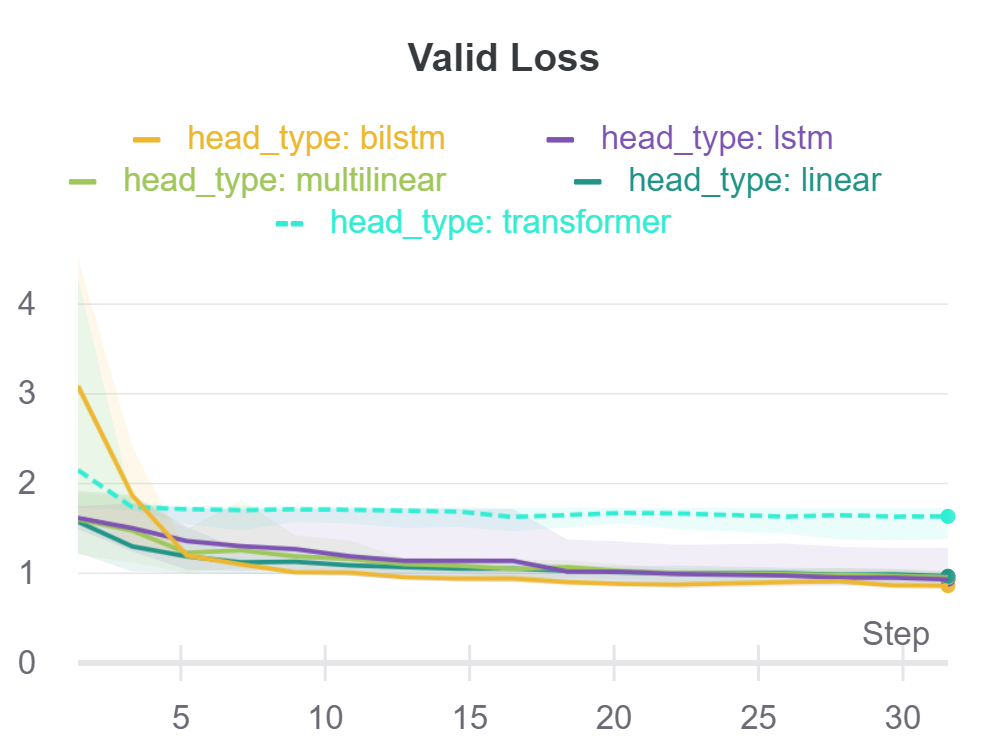}
\includegraphics[width=0.33\linewidth]{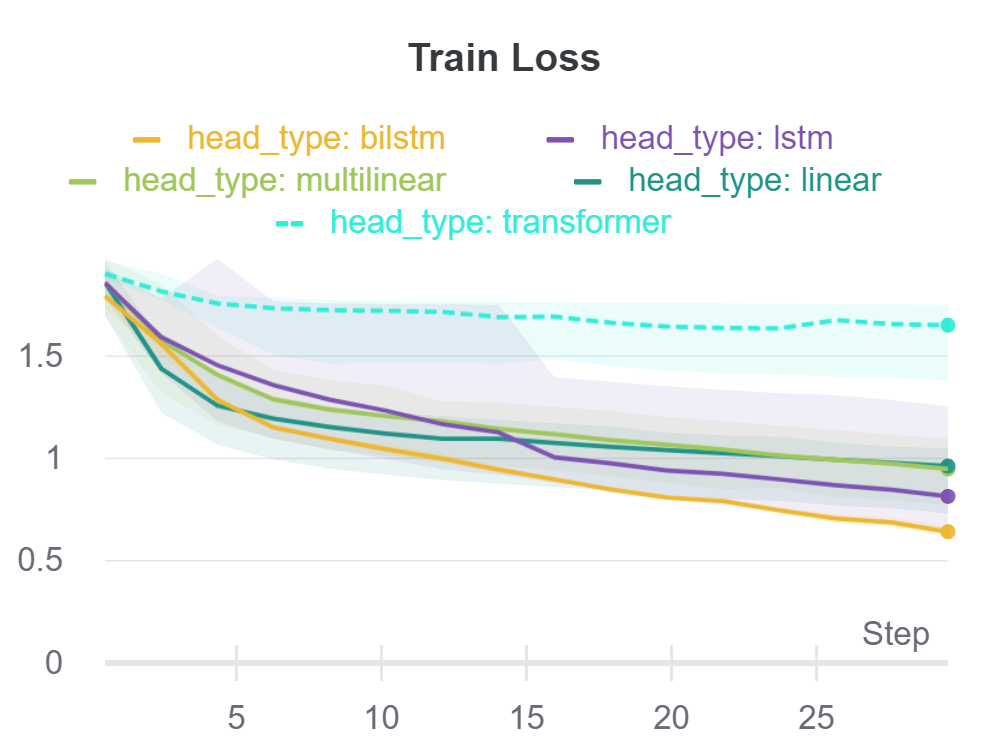}
\includegraphics[width=0.33\linewidth]{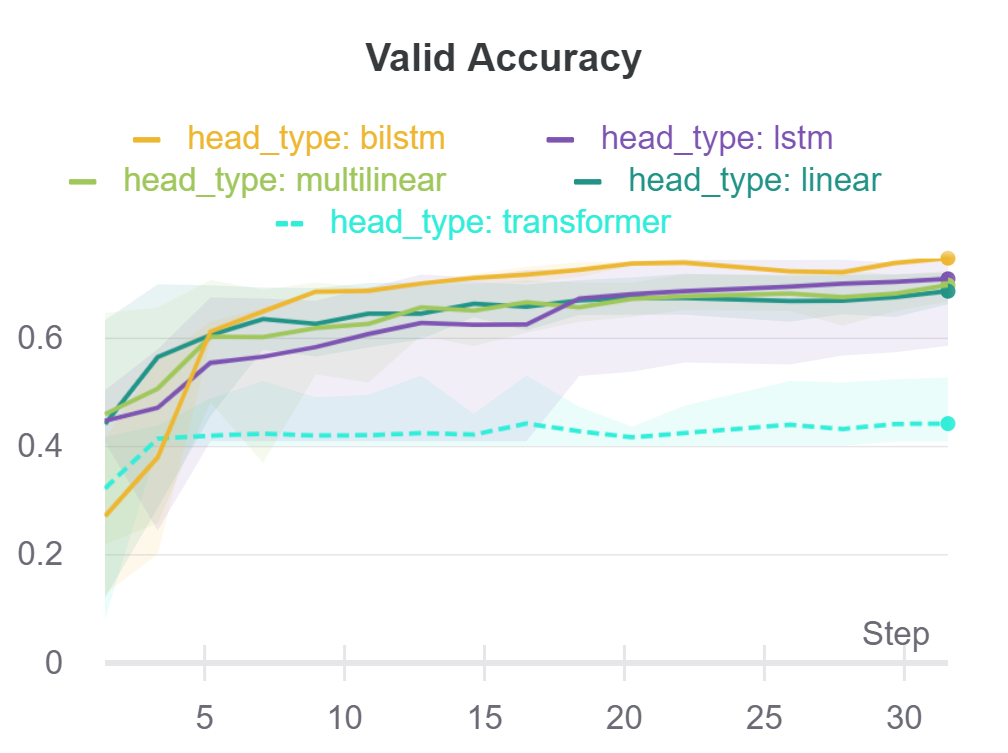}



\end{document}